\documentclass{article}
\usepackage{anyfontsize}

\PassOptionsToPackage{numbers,sort&compress}{natbib}
\usepackage[preprint]{neurips_2026}

\usepackage[utf8]{inputenc}
\usepackage[T1]{fontenc}
\usepackage{hyperref}
\usepackage{url}
\usepackage{booktabs}
\usepackage{amsfonts}
\usepackage{amsmath}
\usepackage{amssymb}
\usepackage{placeins}
\usepackage{nicefrac}
\usepackage{microtype}
\usepackage{xcolor}
\usepackage{graphicx}
\usepackage{float}
\usepackage{multirow}
\usepackage{array}
\usepackage{bm}
\usepackage{xspace}
\usepackage{enumitem}
\usepackage{titlesec}
\usepackage{fontawesome5}

\titlespacing*{\paragraph}{0pt}{0.5ex plus 0.2ex minus 0.1ex}{0.6em}

\newcommand{\modelname}{Flex4DHuman\xspace}
\providecommand{\methodname}{Flex4DHuman}

\DeclareMathOperator{\SE}{SE}

\usepackage[table]{xcolor}
\definecolor{rankone}{RGB}{210,245,218}   
\definecolor{ranktwo}{RGB}{255,242,204}   
\definecolor{rankthree}{RGB}{219,235,255} 

\title{\methodname{}: Flexible Multi-view Video Diffusion for 4D Human Reconstruction}

\author{%
  Jen-Hao Cheng$^{1,2}$ \quad
  Yipeng Wang$^{2,\dagger}$\thanks{$\dagger$ Project lead.} \quad
  Hao Zhang$^{2}$ \quad
  Gengshan Yang$^{2}$ \quad
  Jenq-Neng Hwang$^{1}$ \\
  \normalfont $^1$University of Washington \quad $^2$World Labs \\[3pt]
  \normalfont\small
  \faGlobe~\href{https://andy-cheng.github.io/Flex4DHuman}{\texttt{Project Page}}
  \quad
  \faGithub~\href{https://github.com/Andy-Cheng/Flex4DHuman}{\texttt{Code}}
  \quad
  \raisebox{-0.2ex}{\includegraphics[height=0.95em]{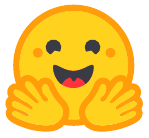}}~\href{https://huggingface.co/datasets/andaba/multi-view_caption}{\texttt{Multi-view Caption Data}}
}

\begin{document}

\maketitle

\begin{figure}[H]
  \centering
  \includegraphics[width=\linewidth]{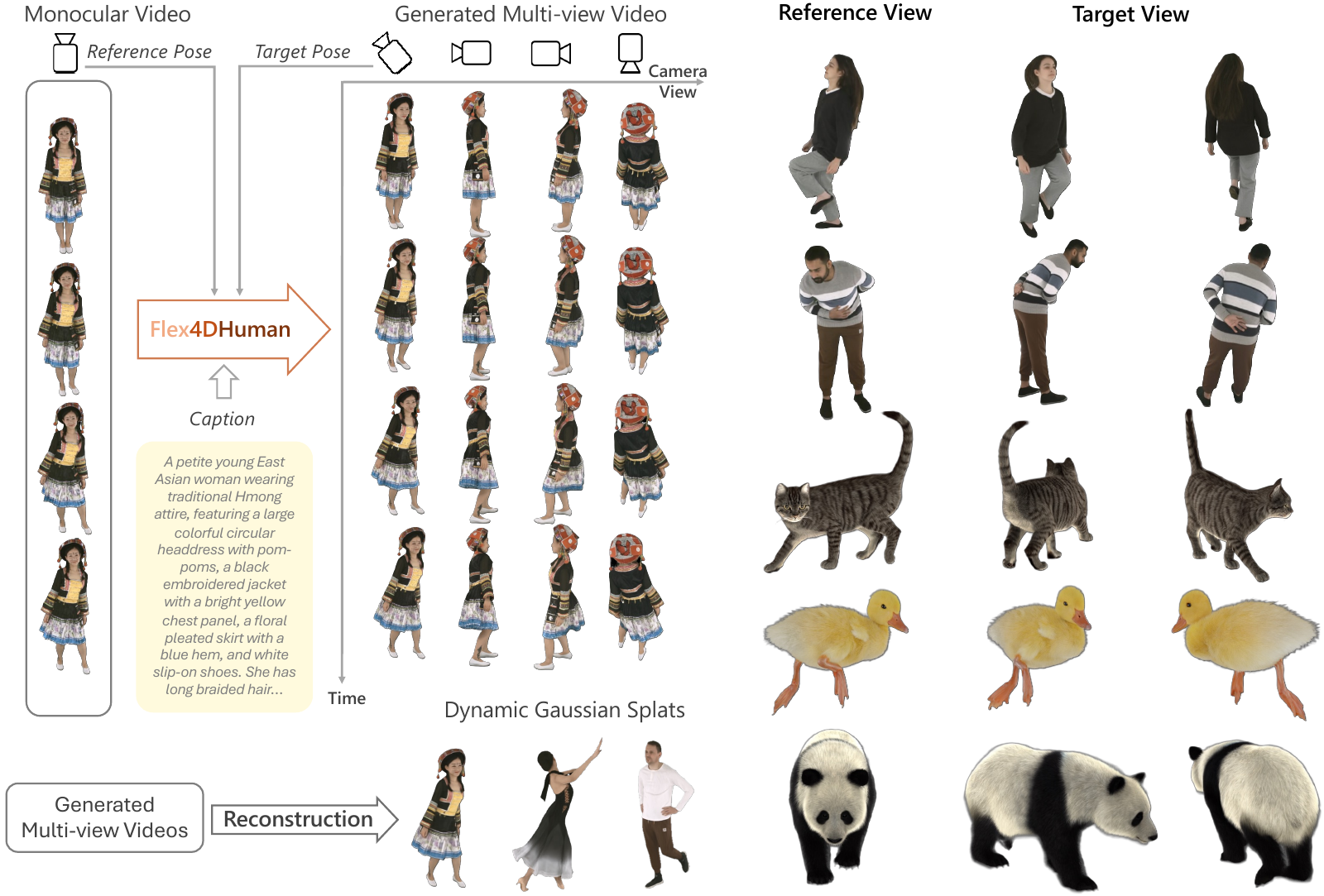}
  \caption{%
    \textbf{\modelname{} turns monocular or sparse-view videos into synchronized dense multi-view videos using only camera-pose and text conditioning.} 
    Given one or more reference-view videos, their camera poses, and target camera poses, the model synthesizes consistent novel-view videos across target views. These generated multi-view videos can be directly used for downstream reconstruction into dynamic Gaussian splats.
    }
  \label{fig:teaser}
\end{figure}

\begin{abstract}
We present Flex4DHuman, a multi-view video diffusion model that transforms a monocular or sparse multi-view video of a dynamic subject into synchronized dense multi-view videos using only relative camera-pose conditioning. 
Unlike prior human-centric methods that rely on skeletons, depth maps, normals, or rendered target-view geometry, Flex4DHuman requires no explicit geometry priors and instead conditions generation through relative camera-pose positional encoding. The generated videos can be directly ingested by downstream reconstruction pipelines to create dynamic 4D Gaussian splats.
Built on Wan 2.1’s 1.3B text-to-video model, Flex4DHuman preserves the backbone architecture and encodes camera and view information through a five-axis positional encoding that extends spatio-temporal RoPE with view indices and continuous SE(3) relative camera geometry. 
A three-stage curriculum progressively trains the model for pose following, flexible reference-to-target view generation, and temporal rollout.
To support temporal rollout, we train with clean historical target-view tokens. We also add multi-view captions to enable test-time text control.
Combined with an off-the-shelf 4D Gaussian Splatting stage, our framework lifts monocular static-camera videos into dynamic 4D Gaussian splats. 
Experiments on DNA-Rendering and ActorsHQ show that Flex4DHuman surpasses prior state-of-the-art methods, while the same formulation generalizes to animal categories after mixed human-animal training. 
These capabilities make Flex4DHuman a practical step toward scalable 4D content creation from casual monocular videos for simulation, gaming, AR/VR, and video re-shooting.
\end{abstract}

\section{Introduction}
\label{sec:introduction}

Recent advances in generative video models have demonstrated strong capabilities in camera-controlled novel-view synthesis~\cite{motionctrl, viewcrafter, ReCamMaster, CameraCtrl2, TrajectoryCrafter, gen3c, BulletTime, UCPE, anyview}. Combined with diffusion priors, these methods point toward free-viewpoint generation of dynamic subjects from sparse visual observations. 
Temporally coherent and cross-view-consistent multi-view videos are especially valuable because they can be directly reconstructed into dynamic 3D representations such as Gaussian splats~\cite{wu20234dgaussian, ftgs}, which can be composed into generated worlds~\cite{marble, hunyuan_world} for downstream applications such as AR/VR, simulation, gaming, and video re-shooting.

A common strategy for multi-view video generation is to condition diffusion models on camera pose, view information, or explicit geometric signals~\cite{mv_adapter, syncammaster, sv4d, 4realv2, diffhuman4d, mvperformer}. For dynamic humans, Diffuman4D~\cite{diffhuman4d} relies on rendered SMPL skeletons in every target view, while MV-Performer~\cite{mvperformer} depends on monocular depth and human normal estimation~\cite{sapiens} to construct partial geometric renderings for target-camera conditioning. Although effective, these approaches inherit the limitations of their geometric inputs: estimation errors propagate into generation, the methods remain tied to human-specific models, and inference is often coupled to fixed camera configurations or reference-view assumptions. Extending such systems to in-the-wild monocular videos or non-human subjects therefore remains challenging.

We instead train a multi-view video diffusion model that does not rely on explicit geometry priors. Starting from Wan~2.1's 1.3B text-to-video DiT~\cite{wan21}, our model synthesizes consistent dense target views from one or more reference-view videos using only relative camera-pose conditioning. It requires no skeleton fitting, depth estimation, normal prediction, or rendered target-view geometry during either training or inference. Our only architectural modification is to replace the self-attention positional encoding with a five-axis formulation that combines spatial coordinates, temporal indices, view-slot indices, and continuous SE(3) camera geometry through PRoPE~\cite{prope}. Because the camera encoding is relative and the view-slot encoding is permutation-invariant, the same model flexibly generalizes across varying numbers of reference views and target-view layouts.

To support appearance synthesis from sparse or monocular inputs, we add multi-view captions during training. Our three-stage curriculum progressively introduces pose following, flexible reference-to-target view conditioning, and temporal rollout. Together, these components enable synchronized multi-view generation with strong temporal and cross-view consistency. Using the generated multi-view videos, we further build a monocular-video-to-4D-Gaussian-splat pipeline that produces dynamic 3D assets from a single static-camera recording without requiring a multi-view capture rig.

We evaluate our method on DNA-Rendering~\cite{dna_rendering} and unseen ActorsHQ~\cite{actorshq}, where it surpasses prior state-of-the-art approaches. On DNA-Rendering, our method outperforms Diffuman4D's GT-skeleton setting by $+1.21$\,dB PSNR / $+0.0037$ SSIM / $-0.0127$ LPIPS, and improves over monocular baselines including Diffuman4D-mono-skeleton and MV-Performer by $+9.32$ and $+8.00$\,dB PSNR, respectively. On zero-shot ActorsHQ, our method outperforms the monocular Diffuman4D setting by $+3.35$\,dB PSNR / $+0.041$ SSIM / $-0.030$ LPIPS. Our experiments further demonstrate robust cross-view consistency under different reference views, monotonic quality improvement with more reference views, and stable temporal rollout. We also fine-tune the same model on DFA animal data~\cite{artemis}, demonstrating generalization beyond humans without architectural changes or human-specific priors.

We summarize our contributions as follows.
\begin{itemize}\itemsep0pt
    \item \textbf{Multi-view video diffusion without explicit geometry priors.}
    We adapt Wan~2.1 into a multi-view video generator that conditions generation through relative camera-pose positional encoding, without using skeletons, depth maps, normals, or rendered target-view geometry.

    \item \textbf{Flexible synchronized generation.}
    Our model supports multiple input and output combinations, including monocular and variable sparse-view inputs, arbitrary target viewpoints, dynamic reference-to-target view conditioning, and temporal rollout.

    \item \textbf{Monocular video to 4D Gaussian splats.}
    We show that the generated synchronized multi-view videos are sufficiently consistent for downstream 4D Gaussian Splatting reconstruction.
\end{itemize}
\section{Related Work}
\label{sec:related_work}

\subsection{Novel view synthesis for dynamic humans}
\label{sec:human_nvs}
Human novel-view synthesis has evolved from subject-specific neural rendering methods toward more generalizable and generative approaches. Early methods such as Neural Body~\cite{peng2021neuralbody}, D-NeRF~\cite{pumarola2021dnerf}, Animatable NeRF~\cite{peng2021animatable}, and Neural Actor~\cite{liu2021neuralactor} demonstrated high-fidelity novel-view rendering of dynamic humans and scenes using neural fields, but typically required subject-specific optimization or densely calibrated multi-view captures. In practice, such data are often collected using specialized multi-camera capture systems and reconstruction pipelines~\cite{joo2017panoptic,xiang2021modeling,luiten2023dynamic}, limiting scalability and accessibility. HumanNeRF~\cite{weng2022humannerf} and NeuMan~\cite{jiang2022neuman} later enabled free-viewpoint rendering from monocular videos, reducing capture requirements while still relying on explicit human-body priors.

Subsequent work focused on improving efficiency and generalization. InstantAvatar~\cite{jiang2023instantavatar} enables rapid avatar reconstruction from monocular videos, while 3DGS-Avatar~\cite{qian20243dgsavatar} and Animatable Gaussians~\cite{li2024animatablegaussians} leverage deformable 3D Gaussian representations for high-quality animatable avatars. GPS-Gaussian~\cite{zheng2024gpsgaussian} further demonstrates real-time, generalizable human novel-view synthesis from sparse observations. While these methods improve reconstruction speed, rendering quality, or cross-subject generalization, they remain reconstruction-based approaches rather than generative models for multi-view video synthesis.

Recent diffusion-based methods have begun to address multi-view video generation. SV4D~\cite{sv4d} generates dynamic 3D content with multi-frame and multi-view consistency, but relies on staged inference-time temporal extension to generate long videos. DiffHuman4D~\cite{diffhuman4d} introduces a spatio-temporal diffusion framework for 4D-consistent human view synthesis from sparse-view videos, but requires accurate human skeletons for both reference and target views. MV-Performer~\cite{mvperformer} incorporates explicit geometric priors and progressively renders target-view conditions through a multi-stage depth, normal, and rendering pipeline before video generation.

In contrast, our approach jointly generates a variable number of target-view videos from a variable number of reference views using only camera-pose conditioning. Unlike prior methods, it does not require skeletons, SMPL models, depth maps, normals, or rendered target-view geometry.

\subsection{Camera conditioning in generative models}
\label{sec:camera_conditioning}

Conditioning generative models on camera geometry is an active area of research. CAT3D~\cite{cat3d} uses raymap conditioning, while CameraCtrl~\cite{cameractrl} and Diffuman4D~\cite{diffhuman4d} employ Pl\"ucker-coordinate camera representations, with Diffuman4D additionally incorporating skeleton-based human priors. SynCamMaster~\cite{syncammaster} encodes camera extrinsics into embeddings for camera-aware cross-view attention. MVDream~\cite{mvdream} uses learned camera embeddings and cross-view attention, while ViewDiff~\cite{viewdiff} combines camera-aware attention with explicit 3D feature projection. Together, these methods span camera-conditioning strategies ranging from spatial ray representations to camera embeddings, attention-based conditioning, and explicit 3D reasoning.

Positional encoding provides a non-parametric alternative. RoPE~\cite{su2024rope} and its extensions encode positional information through rotations rather than learned embeddings, making them inherently length-invariant. PRoPE~\cite{prope} further extends RoPE to continuous $\SE(3)$ transformations, enabling camera geometry to be incorporated directly into the attention mechanism. Building on PRoPE, we combine continuous $\SE(3)$ conditioning with a discrete view-slot encoding based on canonical view order. The discrete view band distinguishes camera identities, while the continuous $\SE(3)$ component generalizes to arbitrary camera layouts. Together, they enable camera-rig-agnostic multi-view generation that is invariant to both the number and spatial arrangement of cameras.
\section{Method}
\label{sec:method}

\subsection{Overview}
\label{ssec:overview}

We build Flex4DHuman on the Wan~2.1 text-to-video DiT~\cite{wan21}
and fine-tune from its 1.3B T2V checkpoint on the DNA-Rendering
dataset~\cite{dna_rendering} with our multi-view captions. Given one
or more reference-view videos, their camera poses, and the desired
target camera poses, the model synthesizes RGB frames at the target
cameras. Generation is performed jointly across views and time within
each temporal chunk, and the chunks are rolled out sequentially to
produce videos longer than the temporal horizon seen during training.
An optional text prompt conditions the subject appearance consistently
across all generated views.

Our key design choice is to preserve the Wan~2.1 backbone and inject
multi-view structure through the self-attention positional encoding.
Specifically, we replace the original spatio-temporal RoPE with a
five-axis formulation that combines spatial coordinates, discrete frame
indices, discrete view-slot indices, and continuous SE(3) camera geometry
via PRoPE~\cite{prope}. This design encodes relative camera geometry
directly inside attention while supporting permutation-invariant view
slots. In addition, we use a clean-conditioning mask to distinguish
observed reference tokens from target tokens. During training, this
allows reference information to propagate bidirectionally across views
and time; during inference, the same mechanism is reused to feed clean
history tokens from previous chunks for temporal rollout. Together,
these components yield a model that extrapolates to denser
view layouts and longer temporal sequences than those seen during
training.

We describe the architecture and camera-conditioned positional encoding
in \S~\ref{ssec:arch}, followed by the training curriculum in
\S~\ref{ssec:training}.


\subsection{\modelname{} architecture}
\label{ssec:arch}
\begin{figure*}[t]
  \centering
  \includegraphics[width=1\textwidth]{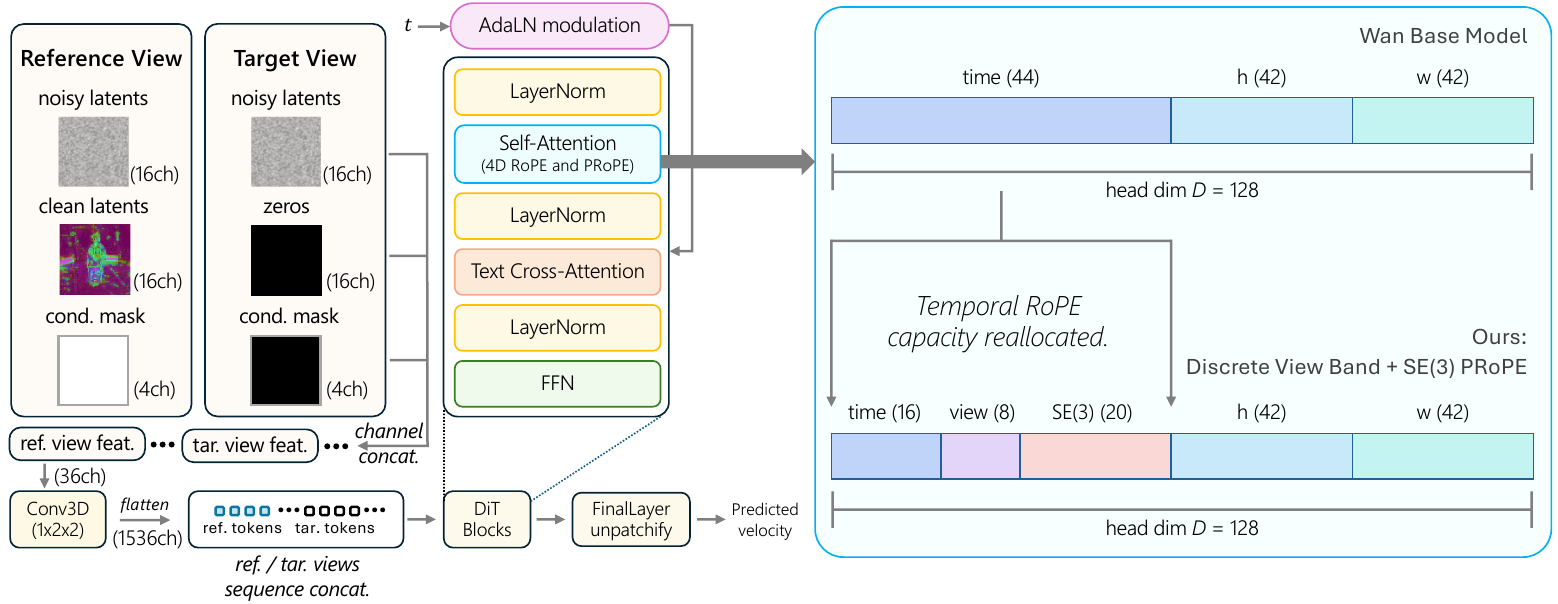}
  \caption{\textbf{\modelname{} architecture and projective positional encoding.} Reference and target views are represented using a 36-channel feature layout consisting of 16-channel noisy latents, 16-channel clean latents (set to zero for target views), and a 4-channel binary mask indicating reference views. After a $\mathrm{Conv3D}(1{\times}2{\times}2)$ spatial patch downsampler, the per-view features are flattened into token sequences and processed by the DiT backbone. 
  Self-attention employs a projective positional encoding that augments RoPE with view and camera-geometry information. Whereas Wan allocates the attention head dimension $D{=}128$ across temporal and spatial bands as $(D_t,D_h,D_w)=(44,42,42)$, we reallocate the temporal RoPE capacity into separate frame, view, and continuous SE(3) geometry bands, $(D_{\text{frame}},D_{\text{view}},D_{\text{SE(3)}})=(16,8,20)$. This allocation enables the attention layers to jointly encode temporal position, view identity, and relative camera geometry while preserving the original spatial encoding capacity.
}
  \label{fig:arch_rope}
\end{figure*}

Our architecture, shown in Fig.~\ref{fig:arch_rope}, builds upon the Wan~2.1 1.3B T2V DiT backbone, with the only architectural modification being the positional encoding used in self-attention. Tokens are packed in \emph{view-major} order, allowing each token to be indexed by both its view and temporal position.

Reference and target views are represented using a shared 36-channel input layout (Fig.~\ref{fig:arch_rope}) comprising 16 noisy latent channels, 16 conditioning latent channels, and a 4-channel binary conditioning mask inherited from Wan~2.1 I2V. 
Reference views populate the conditioning channels with encoded reference-view latents and all-one masks, whereas target views use zeros. 
Because the original Wan~2.1 T2V checkpoint expects 16 input channels, we expand the input projection to 36 channels by copying the pretrained weights for the original channels and zero-initializing the newly introduced parameters. This representation allows information to propagate jointly across views and time within a single attention operation.

To encode camera geometry, we build on PRoPE, which extends rotary positional encoding to SE(3) camera transformations. For each token, a dedicated slice of the query and key vectors is transformed according to its camera pose:
\begin{equation}
Q^\text{SE(3)}_{i}
\leftarrow
\mathbf{T}^{\!\top}_{i}\,
Q^\text{SE(3)}_{i},
\qquad
K^\text{SE(3)}_{j}
\leftarrow
\mathbf{T}^{-1}_{j}\,
K^\text{SE(3)}_{j},
\end{equation}
where $Q^\text{SE(3)}_{i}$ and $K^\text{SE(3)}_{j}$ are the query and key sub-vectors allocated to the SE(3) band, and $\mathbf{T}_{i}, \mathbf{T}_{j}\in\mathrm{SE}(3)$ denote the camera poses associated with tokens $i$ and $j$. Applying $\mathbf{T}^{\!\top}_{i}$ to queries and $\mathbf{T}^{-1}_{j}$ to keys makes the resulting attention depend on the relative camera transformation between tokens. This formulation introduces no additional input channels or learnable camera parameters and naturally generalizes to arbitrary reference and target-view configurations.

Wan~2.1 allocates the RoPE dimensions of each attention head as
$(D_t,D_h,D_w)=(44,42,42)$ for head dimension $D=128$. We repartition the temporal band into separate time, view, and SE(3) geometry sub-bands:
\begin{equation}
(D_\text{time}, D_\text{view}, D_\text{SE(3)}, D_h, D_w)
=
(16, 8, 20, 42, 42).
\end{equation}
As illustrated in Fig.~\ref{fig:arch_rope}, the view band provides a discrete identifier for camera viewpoints, while the SE(3) band encodes relative camera geometry directly within attention. For numerical stability, camera poses are normalized on a per-sequence basis by expressing all cameras relative to the first camera and scaling translations to unit distance. Because the SE(3) component reuses RoPE dimensions originally allocated to temporal encoding, the pretrained positional prior learned from motion in Wan~2.1 can be naturally extended to multi-view camera geometry without introducing additional parameters.

\subsection{Training}
\label{ssec:training}

\paragraph{Objective.}
We use the same flow-matching objective as Wan~2.1. Given a clean latent
$\mathbf{x}_1$, Gaussian noise $\mathbf{x}_0 \sim \mathcal{N}(0,I)$, and
a time step $t \in [0,1]$ sampled from the logit-normal schedule, we
construct the linear interpolant
$\mathbf{x}_t=(1-t)\mathbf{x}_0+t\mathbf{x}_1$. The model predicts the
velocity $\mathbf{v}_\theta(\mathbf{x}_t,t)$ and minimizes
\begin{equation}
    \mathcal{L}
    =
    \mathbb{E}
    \left[
    \left\|
    \mathbf{v}_\theta(\mathbf{x}_t,t)
    -
    (\mathbf{x}_1-\mathbf{x}_0)
    \right\|_2^2
    \right],
\end{equation}
with the loss averaged uniformly across all tokens. Clean reference and
history tokens are provided through the conditioning channels described
in \S~\ref{ssec:arch}, so the model learns to preserve observed tokens
and denoise unknown target tokens. We drop the text condition to a zero
embedding with probability $0.1$ for classifier-free guidance.

\paragraph{Curriculum.}
\begin{figure*}[t]
  \centering
  \includegraphics[width=1\textwidth]{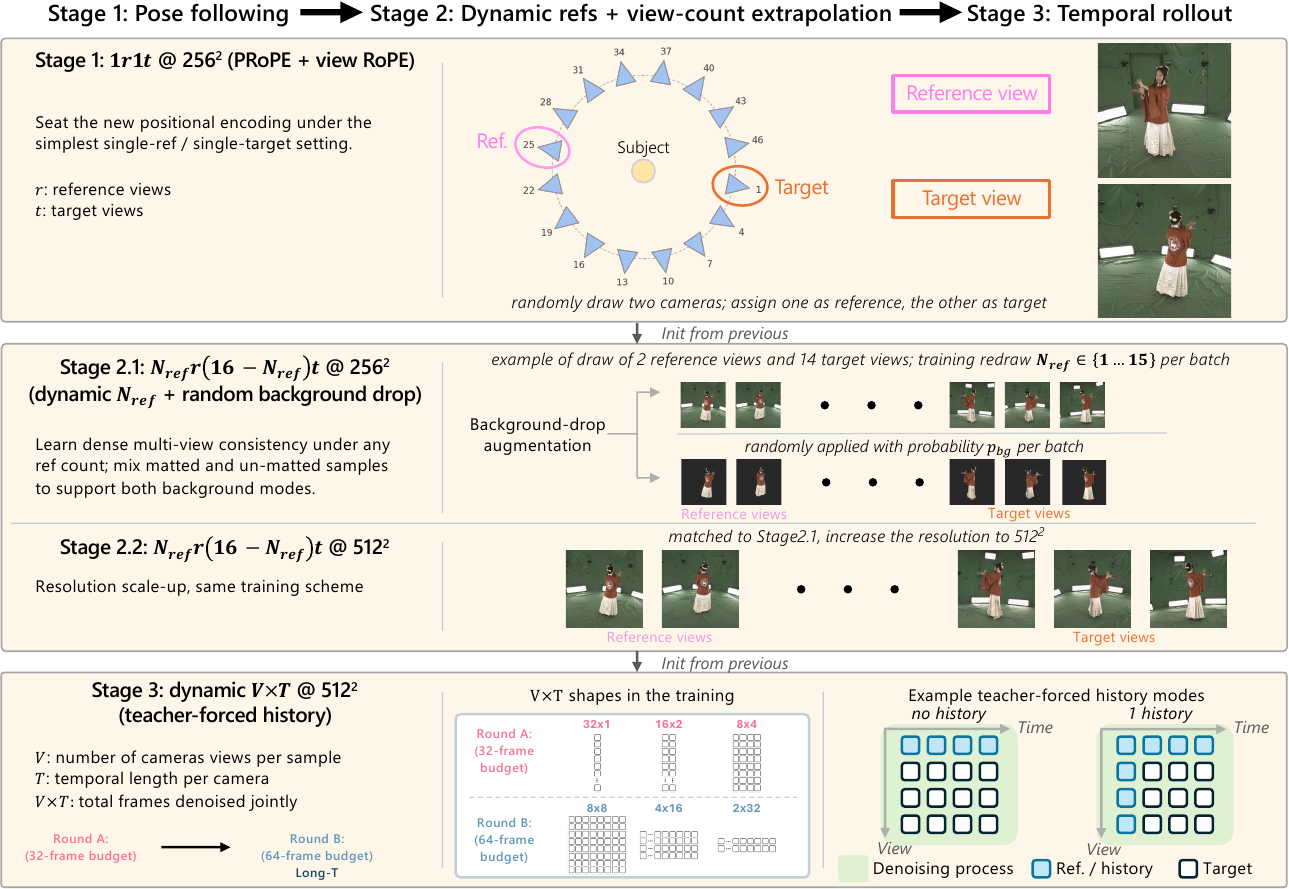}
\caption{
\textbf{Training curriculum and temporal rollout.}
\textbf{Three-stage training curriculum.}
Stage~1 adapts the pretrained backbone to the new camera-aware positional encoding in a single-reference single-target setting. 
Stage~2 introduces dynamic reference-view sampling with random background-drop augmentation, enabling synchronized multi-view generation under variable reference-to-target view configurations.
Stage~3 trains on dynamic view-time layouts with teacher-forced history conditioning, enabling temporal rollout beyond the training window.
}
  \label{fig:curriculum}
\end{figure*}

We train \modelname{} using the three-stage curriculum shown in
Fig.~\ref{fig:curriculum}. Each stage initializes from the previous
checkpoint and introduces an additional axis of difficulty.

\textbf{Stage~1} uses the simplest single-reference single-target setting
($1r1t$, $T{=}1$) at $256^2$, where $r/t$ denote reference/target views
and $T$ denotes the number of frames per view. 
This allows the pretrained Wan~2.1 backbone to adapt to the new camera- and view-aware positional encoding.

\textbf{Stage~2} introduces dynamic reference-view sampling with
$N_\text{ref} \in \{1,\dots,15\}$ under a fixed total view count
$V{=}16$ and $T{=}1$, where $N_\text{ref}$ is the number of reference
views and $V$ is the total number of views. This enables synchronized
multi-view generation across varying reference-to-target view
configurations. This stage also applies random background-drop augmentation and is trained first at $256^2$ before being scaled to $512^2$ for finer detail.

\textbf{Stage~3} extends training to dynamic temporal windows with teacher-forced
history conditioning at $512^2$, enabling temporal rollout over long
sequences. We multitask several $V{\times}T$ configurations under a
shared token budget, including $\{32{\times}1, 16{\times}2, 8{\times}4\}$
and $\{8{\times}8, 4{\times}16, 2{\times}32\}$, allowing one checkpoint
to generalize across view counts and temporal lengths. From Stage~2
onward, we randomly remove the background, enabling both foreground-only and
full-background generation.

\paragraph{Inference-time view and temporal extrapolation.}
At inference time, we extrapolate beyond the training distribution along
both the view and temporal dimensions. For view extrapolation, the model
synthesizes substantially more target views than observed during training.

\begin{figure*}[t]
  \centering
  \includegraphics[width=0.8\textwidth]{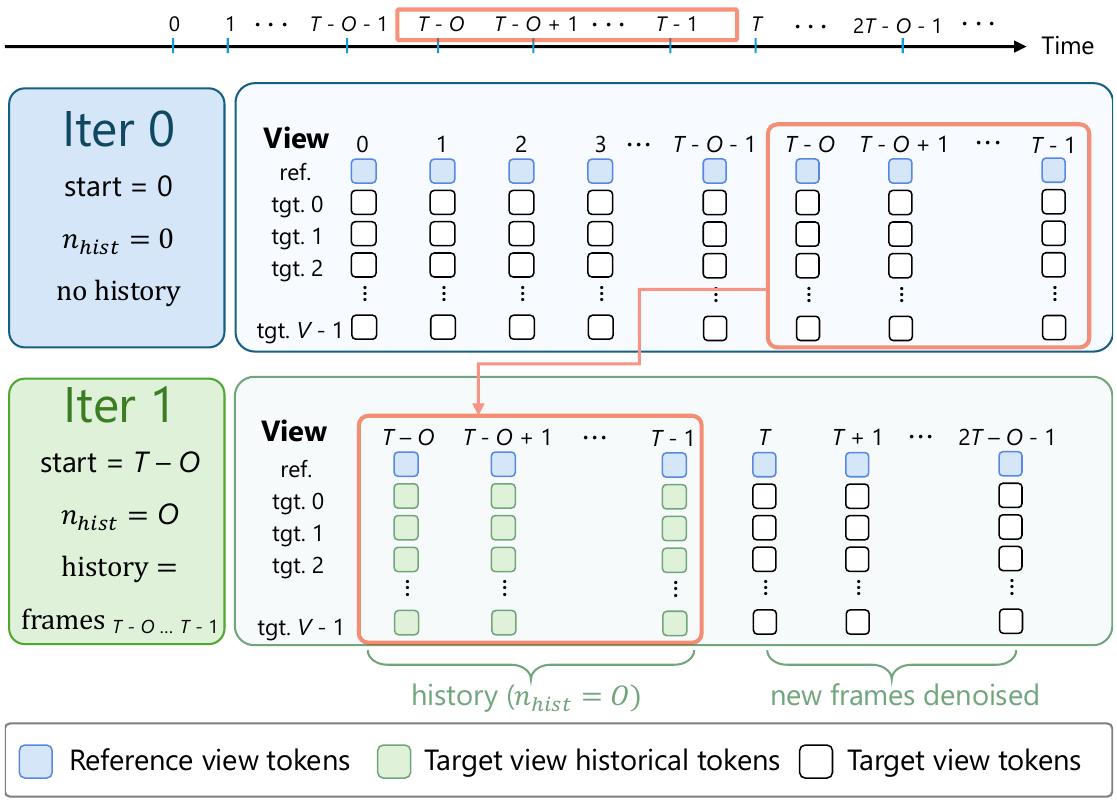}
\caption{
\textbf{Temporal rollout.}
Each iteration denoises a $T$-frame chunk across all views. 
The first iteration uses only the reference-view tokens as clean conditioning. 
Subsequent iterations advance by $T-\mathcal{O}$ frames and reuse the $\mathcal{O}$ overlapping predictions from the previous chunk as clean history tokens for all target views, enabling long-horizon synchronized multi-view generation.
}
  \label{fig:rollout}
\end{figure*}

For temporal extrapolation, we use the chunked rollout strategy shown in
Fig.~\ref{fig:rollout}, with an overlap of $\mathcal{O}$ frames
between consecutive chunks. The overlapping predictions from the previous
chunk are reused as clean history conditioning for the next chunk,
enabling long-horizon synchronized multi-view generation.

\modelname{} can lift a monocular video into a complete dynamic 3D asset. Given a static-camera recording and user-specified target cameras, the model first generates synchronized multi-view videos, then we fit FreeTimeGS~\cite{ftgs} to reconstruct dynamic Gaussian splats.

\section{Multi-view Captioning}
\label{sec:caption}

Our data consists of three multi-view video datasets:
DNA-Rendering~\cite{dna_rendering}, using the Diffuman4D~\cite{diffhuman4d}
processed release, which contains $1{,}038$ human performance sequences from
$548$ identities captured with a $48$-camera rig;
ActorsHQ~\cite{actorshq}, containing $14$ sequences from $8$ actors captured
with $160$ cameras; and the Dynamic Furry Animals (DFA) dataset from
Artemis~\cite{artemis}, consisting of $23$ sequences spanning $9$ animal
species rendered from $36$--$72$ camera viewpoints
(Table~\ref{tab:caption_data}). 

\begin{table}[t]
  \centering
  \small
  \resizebox{\linewidth}{!}{%
  \begin{tabular}{l l rrr c r r rrr}
    \toprule
    Source & Subject & \#IDs & \#Seqs & \#Views & Resolution & FPS &
    \#Frames & Window & Captions & Average Words \\
    \midrule
    DNA-Rendering~\cite{dna_rendering,diffhuman4d} & Human & 548 &
    1{,}038 & 48 & $1024 \times 1024$ & 15 & 229.2k &
    10\,fr\,/\,0.7\,s & 23{,}410 & 268 \\
    ActorsHQ~\cite{actorshq} & Human & 8 & 14 & 160 &
    $747 \times 1022$ & 25 & 31.2k &
    20\,fr\,/\,0.8\,s & 1{,}566 & 269 \\
    DFA~\cite{artemis} & Animal & 9 & 23 & 36\,/\,72 &
    $1920 \times 1080$ & 30 & 4.0k &
    60\,fr\,/\,2.0\,s & 55 & 238 \\
    \midrule
    Total & & 565 & 1{,}075 & & & & 264.4k & & 25{,}031 & 268 \\
    \bottomrule
  \end{tabular}}
  \caption{\textbf{Multi-view caption corpus.}  \#IDs counts
    distinct subjects, \#Frames counts per-view frames, and Window
    is the non-overlapping captioning window (frames / seconds).}
  \label{tab:caption_data}
\end{table}

We generate dense natural-language
descriptions for every sequence using Gemini~3 Flash~\cite{gemini3},
enabling text conditioning during both training and inference.

Each sequence is divided into non-overlapping temporal windows
($10$, $20$, and $60$ frames for DNA-Rendering, ActorsHQ, and DFA,
respectively; Table~\ref{tab:caption_data}). Within each window, we
uniformly sample frames and construct a $2{\times}2$ image grid from
four approximately orthogonal viewpoints (front, back, left, and right)
centered on the foreground subject, with the background masked out.
The resulting grid-frame sequence is provided to Gemini together with a
prompt requesting a detailed description of the subject's appearance,
including body shape, hair or fur characteristics, clothing,
accessories, and any visible text or logos. For animals, the prompt
additionally requests a summary of observed behaviors, such as gait,
head posture, and tail movement. This process produces $25{,}031$
captions with an average length of $268$ words.

We intentionally focus human captions on \emph{appearance} rather than
fine-grained motion. During pilot studies, we found that generated
motion descriptions frequently misidentify motion direction (e.g.,
left/right or forward/backward) when subjects are viewed from
non-canonical camera angles. Such errors introduce noisy supervision
that can hinder training. By emphasizing appearance attributes, we
obtain more reliable and temporally stable conditioning signals.
For animals, high-level behavioral descriptions are retained,
as actions such as walking, trotting, running, or tail carriage are
typically defined in body-centric coordinates and remain consistent
across viewpoints.

During training, each sampled clip is paired with the caption
corresponding to the temporal window containing its start frame.
Consequently, different clips from the same sequence are exposed to
different textual descriptions over time, increasing caption diversity
and reducing overfitting to a single static description.
\providecommand{\methodname}{Flex4DHuman}

\section{Experiments}
\label{sec:experiments}

We evaluate \modelname{} on three benchmarks. On DNA-Rendering~\cite{dna_rendering}
and DFA~\cite{artemis}, we directly compare generated novel views against
ground-truth target frames. On the held-out ActorsHQ~\cite{actorshq} capture
rig, we fit the generated multi-view videos with FreeTimeGS~\cite{ftgs},
re-render the reconstructed 4D Gaussian splats at the ActorsHQ ground-truth
cameras, and evaluate the re-rendered results. All metrics are computed on the
subject foreground using PSNR, SSIM, and LPIPS.

\subsection{Training and inference setup}
\label{ssec:exp_setup}

We train \modelname{} on DNA-Rendering using the three-stage curriculum
described in \S~\ref{ssec:training} with $32{\times}$H100 GPUs. Stage~1 is
trained for $30$k iterations, Stages~2.1 and~2.2 for $30$k iterations each, and
Stage~3 for $15$k iterations. At inference time, we use $40$ denoising steps
with a text classifier-free guidance weight of $3.0$.

\paragraph{Comparison methods.}
We compare against two multi-view human diffusion baselines, providing
each method with the same single reference video.

\emph{Diffuman4D-GT-skeleton}~\cite{diffhuman4d} conditions on SMPL
skeletons triangulated from all reference and target views in the
calibrated multi-view rig, a best-case setting unavailable in real
single-reference deployments.

\emph{Diffuman4D-mono-skeleton} replaces this oracle input with a
skeleton lifted from the single reference using Sapiens-Pose and
Sapiens-Depth, anchored to the scene's ground-truth hip location and
body height.

\emph{MV-Performer}~\cite{mvperformer} estimates monocular depth and
human normals from the reference view, splats the resulting point cloud
into target views, and refines the renderings with a diffusion model.

\subsection{Results on DNA-Rendering}
\label{ssec:dna_main}

We evaluate on the 16-scene Diffuman4D test split using $1$ reference
view and $47$ target views. Since the baselines operate on
foreground-only inputs, we report \modelname{} in two settings:
\emph{foreground-only}, using a matted reference, and \emph{whole-scene},
using the full-RGB reference.

\begin{table}[h]
  \centering
  \setlength{\tabcolsep}{0pt}
  \caption{\textbf{Results on DNA-Rendering.}
  Results are evaluated on the 16-scene Diffuman4D test split in a
  $1$-reference, $47$-target setting. We report per-frame
  foreground-only metrics against ground-truth target views. Best results
  are shown in \textbf{bold}.}
  \label{tab:dna_main}
  \begin{tabular*}{\linewidth}{@{\extracolsep{\fill}} l c c c @{}}
    \toprule
    Method & PSNR $\uparrow$ & SSIM $\uparrow$ & LPIPS $\downarrow$ \\
    \midrule
    \textsf{MV-Performer~\cite{mvperformer}}                     & 17.44 & 0.7204 & 0.2697 \\
    \textsf{Diffuman4D-mono-skeleton~\cite{diffhuman4d}}         & 16.12 & 0.8760 & 0.1580 \\
    \textsf{Diffuman4D-GT-skeleton~\cite{diffhuman4d}}           & 24.23 & 0.9479 & 0.0744 \\
    \textsf{\modelname{}-unmatted (ours)}                        & 25.27 & 0.9268 & 0.0977 \\
    \textbf{\textsf{\modelname{}-fg (ours)}}                     & \textbf{25.44} & \textbf{0.9516} & \textbf{0.0617} \\
    \bottomrule
  \end{tabular*}
\end{table}

As shown in Table~\ref{tab:dna_main}, \modelname{}-fg outperforms the
matched single-reference baselines by $+9.32$\,dB PSNR over
Diffuman4D-mono-skeleton and $+8.00$\,dB PSNR over MV-Performer. It also
surpasses Diffuman4D-GT-skeleton by $+1.21$\,dB PSNR, $+0.0037$ SSIM,
and $-0.0127$ LPIPS, despite using less information at inference. In the remaining DNA-Rendering experiments, we compare
against the strongest Diffuman4D variant, Diffuman4D-GT-skeleton, in the
foreground-only setting.

\begin{figure*}[h]
  \centering
  \includegraphics[width=\linewidth]{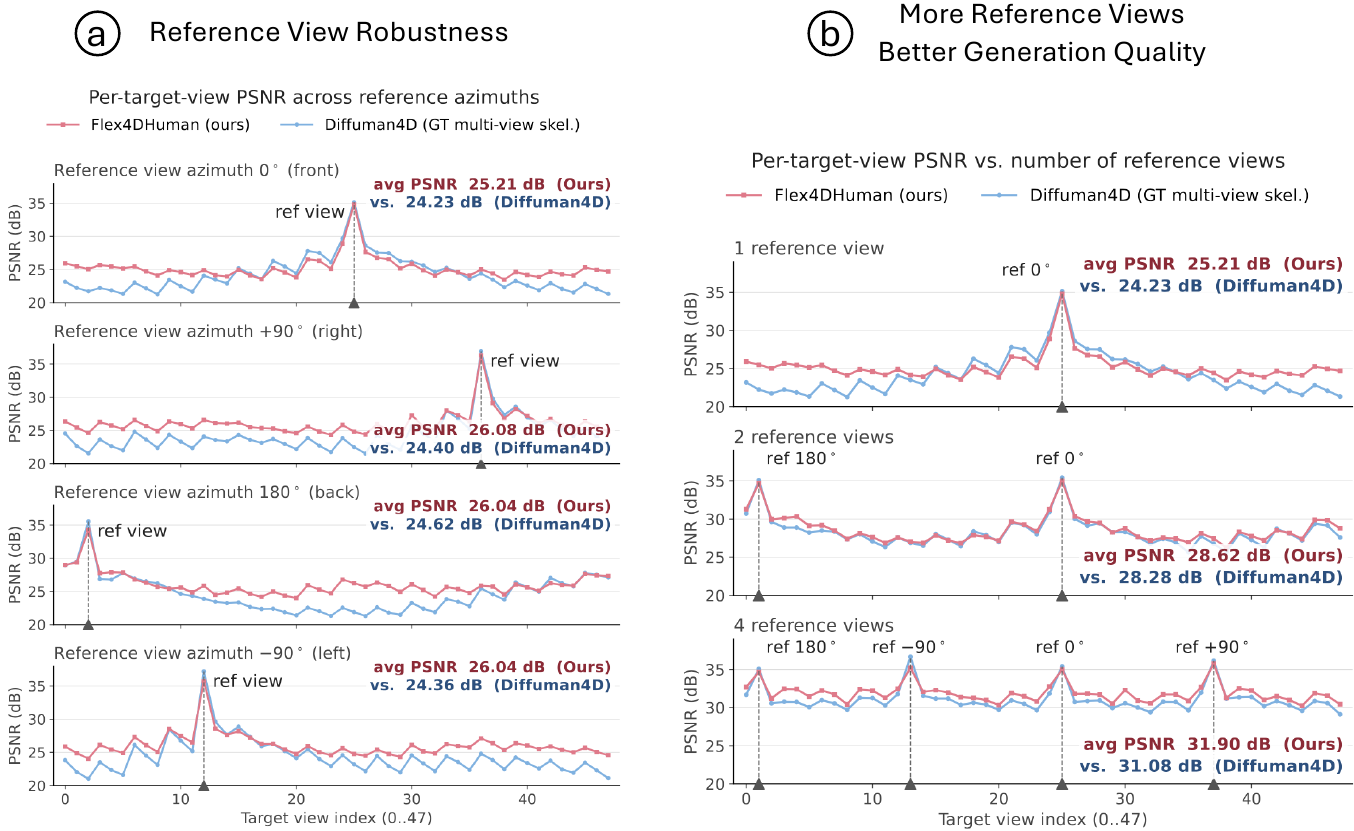}
  \caption{
  \textbf{Analysis on the DNA-Rendering test set.}
  \textbf{(a) Reference-view robustness.}
  We evaluate per-target-view PSNR under four cardinal reference azimuths:
  front $0^\circ$, right $+90^\circ$, back $180^\circ$, and left $-90^\circ$.
  Each panel fixes one reference view and plots the 16-scene PSNR for
  \modelname{} and Diffuman4D-GT-skeleton; vertical markers indicate the
  reference-view column.
  \textbf{(b) Reference-view scaling.}
  We vary the number of reference views from $1$ to $2$ and $4$ while keeping
  the same target-view protocol. Vertical markers indicate the reference views.
  }
  \label{fig:dna_analysis}
\end{figure*}

We further analyze \modelname{} on the DNA-Rendering test set along three
axes: reference-view robustness, reference-view scaling, and temporal
rollout, as shown in Fig.~\ref{fig:dna_analysis}.

\paragraph{Reference-view robustness.}
A relative camera encoding should produce comparable quality regardless of
the input reference azimuth. We therefore select one reference view from
each cardinal direction: front $0^\circ$, right $+90^\circ$, back
$180^\circ$, and left $-90^\circ$, and evaluate the first $40$ frames per
scene with $47$ target views across the same $16$ scenes. As shown in
Fig.~\ref{fig:dna_analysis}\,(a), \modelname{} maintains a cross-azimuth
mean between $25.2$ and $26.1$\,dB, with less than $1$\,dB variation across
reference views. This indicates that the model is robust to the choice of
monocular reference view, benefiting from dynamic reference-view training
and the relative camera encoding.

\paragraph{Reference-view scaling.}
We next vary the number of reference views under the same first-$40$-frame
DNA-Rendering protocol: $1$ reference at $0^\circ$, $2$ references at
$\{0^\circ,180^\circ\}$, and $4$ references at
$\{0^\circ,\pm90^\circ,180^\circ\}$. As shown in
Fig.~\ref{fig:dna_analysis}\,(b), the over-target PSNR improves
monotonically from $25.21$ to $28.62$ and $31.90$\,dB as the number of
reference views increases. The same checkpoint absorbs additional reference
evidence without retraining, confirming the effectiveness of the
dynamic-$N_\text{ref}$ training schedule.

\paragraph{Temporal rollout.}
Finally, we evaluate Stage~3's teacher-forced temporal training for
chunked rollout. We compare two settings with overlap $\mathcal{O}{=}1$:
$T{=}4$ rollout with $14$ iterations and $T{=}16$ rollout with $3$
iterations, both evaluated over the same $42$-frame window with reference
camera~$25$ and $47$ target views.

\begin{figure*}[h]
  \centering
  \includegraphics[width=0.8\linewidth]{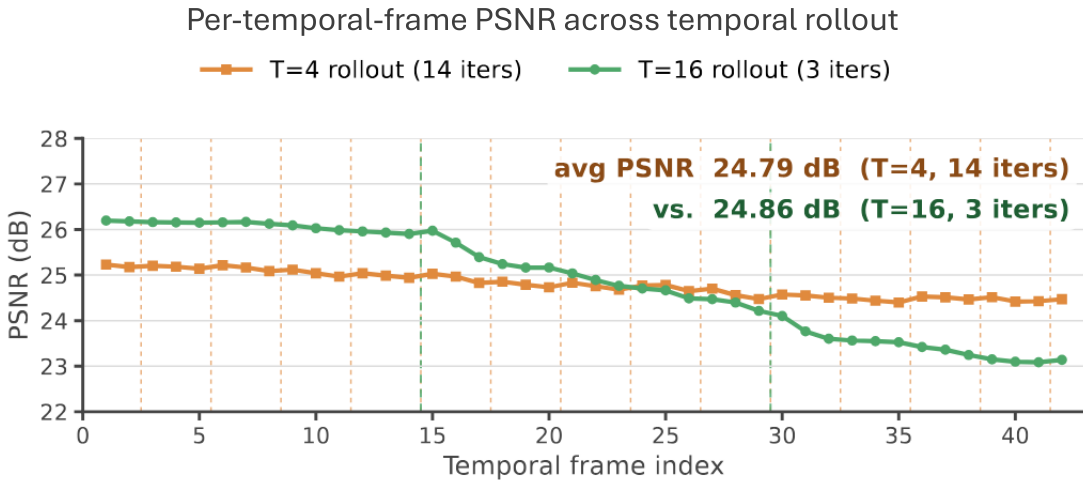}
  \caption{
  \textbf{Temporal rollout.}
  We compare chunked rollout with $T{=}4$ and $T{=}16$ under overlap
  $\mathcal{O}{=}1$ over the same 42-frame window. Dotted vertical lines mark
  chunk boundaries.
  }
  \label{fig:rollout_exp}
\end{figure*}

As shown in
Fig.~\ref{fig:rollout_exp}, the two settings achieve similar
PSNR, $24.79$\,dB for $T{=}4$ and $24.86$\,dB for $T{=}16$, despite
$T{=}4$ requiring more rollout iterations. This suggests that the
teacher-forced history conditioning supports stable long-horizon rollout,
while short chunks provide a memory-efficient operating point for dense
multi-view generation.

\begin{figure*}[!h]
  \centering
  \includegraphics[width=\textwidth,height=0.89\textheight,keepaspectratio]{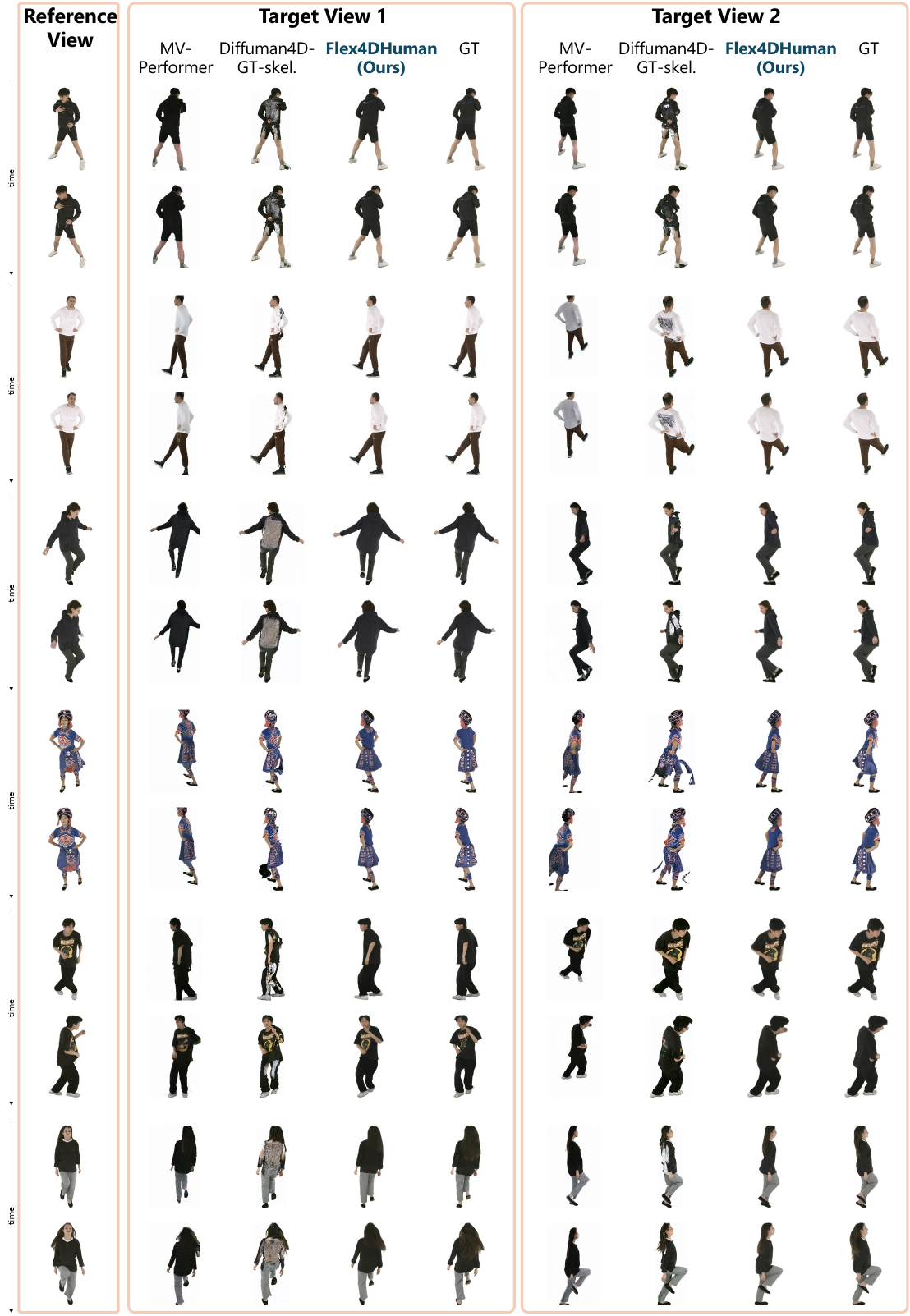}
  \caption{\textbf{Qualitative comparison}
    For each scene a reference view (left
    column) is shown alongside two target views;
    within each target column we compare
    MV-Performer~\cite{mvperformer},
    Diffuman4D-GT-skeleton~\cite{diffhuman4d}, \modelname{}-fg
    (ours), and the ground truth.}
  \label{fig:dna_qual}
\end{figure*}

Figure~\ref{fig:dna_qual} shows qualitative comparisons on
DNA-Rendering.

\subsection{Zero-shot generalization on ActorsHQ}
\label{ssec:actorshq}

ActorsHQ is held out from training and uses a different camera rig from
DNA-Rendering. To evaluate zero-shot 4D reconstruction, we first generate
synchronized multi-view videos on a fixed synthetic target rig, fit
FreeTimeGS~\cite{ftgs} to the generated views, and re-render the
reconstructed 4D Gaussian splats at the ActorsHQ ground-truth cameras.
We evaluate $14$ sequences, including the $12$ sequences used by
Diffuman4D and two additional sequences,
\textsf{Actor04\_Sequence2} and \textsf{Actor06\_Sequence2}, over
$200$ frames per sequence. We compare against
the Diffuman4D-mono-skeleton setting, which follows the same
single-reference constraint as \modelname{}.

\begin{table}[h]
  \centering
  \setlength{\tabcolsep}{0pt}
  \caption{\textbf{Zero-shot generalization on ActorsHQ.}
  We evaluate $14$ sequences with $200$ frames each. Generated views are
  fit with FreeTimeGS and re-rendered at the ground-truth ActorsHQ
  cameras. Best results are shown in \textbf{bold}.}
  \label{tab:actorshq}
  \begin{tabular*}{\linewidth}{@{\extracolsep{\fill}} l c c c @{}}
    \toprule
    Method & PSNR $\uparrow$ & SSIM $\uparrow$ & LPIPS $\downarrow$ \\
    \midrule
    \textsf{Diffuman4D-mono-skeleton~\cite{diffhuman4d}} 
      & 17.97 & 0.815 & 0.307 \\
    \textbf{\textsf{\modelname{}-fg (ours)}}             
      & \textbf{21.32} & \textbf{0.856} & \textbf{0.277} \\
    \bottomrule
  \end{tabular*}
\end{table}

As shown in Table~\ref{tab:actorshq}, \modelname{} outperforms
Diffuman4D-mono-skeleton by $+3.35$\,dB PSNR, $+0.041$ SSIM, and
$-0.030$ LPIPS. The qualitative gap is most visible on rear-facing
target views, where monocular human pose estimation is less reliable. 
In contrast, \modelname{} does not rely on explicit human geometry priors,
making it more robust to cross-rig camera shifts.

\subsection{Beyond humans: multi-view animals}
\label{ssec:exp_dfa}

To evaluate generalization beyond humans, we fine-tune the Stage~3
checkpoint on Dynamic Furry Animal dataset (DFA)~\cite{artemis}. We consider two regimes: \emph{within-animal},
where all species appear in training but test clips are held out, and
\emph{cross-animal}, where the target species are excluded from training.
Generated views are directly evaluated against ground-truth target views
at $T{=}1$ with frame stride $8$.

\begin{table}[h]
  \centering
  \setlength{\tabcolsep}{0pt}
  \caption{\textbf{Animal multi-view generation on DFA.}
  We report aggregate per-frame foreground-only metrics against
  ground-truth target views under within-animal and cross-animal
  evaluation.}
  \label{tab:dfa}
  \begin{tabular*}{\linewidth}{@{\extracolsep{\fill}} l c c c @{}}
    \toprule
    Regime & PSNR $\uparrow$ & SSIM $\uparrow$ & LPIPS $\downarrow$ \\
    \midrule
    Within-animal mean ($n{=}6$) & 22.16 & 0.9079 & 0.0925 \\
    Cross-animal mean ($n{=}3$)  & 20.32 & 0.8757 & 0.1097 \\
    \bottomrule
  \end{tabular*}
\end{table}

As shown in Table~\ref{tab:dfa}, \modelname{} reaches $22.16$\,dB PSNR
and $0.9079$ SSIM in the within-animal setting. In the cross-animal
setting, performance decreases by only about $1.8$\,dB PSNR on average,
showing that the same formulation transfers beyond humans with a small
fine-tuning budget and without human-specific geometry priors.

\section{Applications}
\label{sec:applications}

\begin{figure*}[t]
  \centering
  \includegraphics[width=\textwidth]{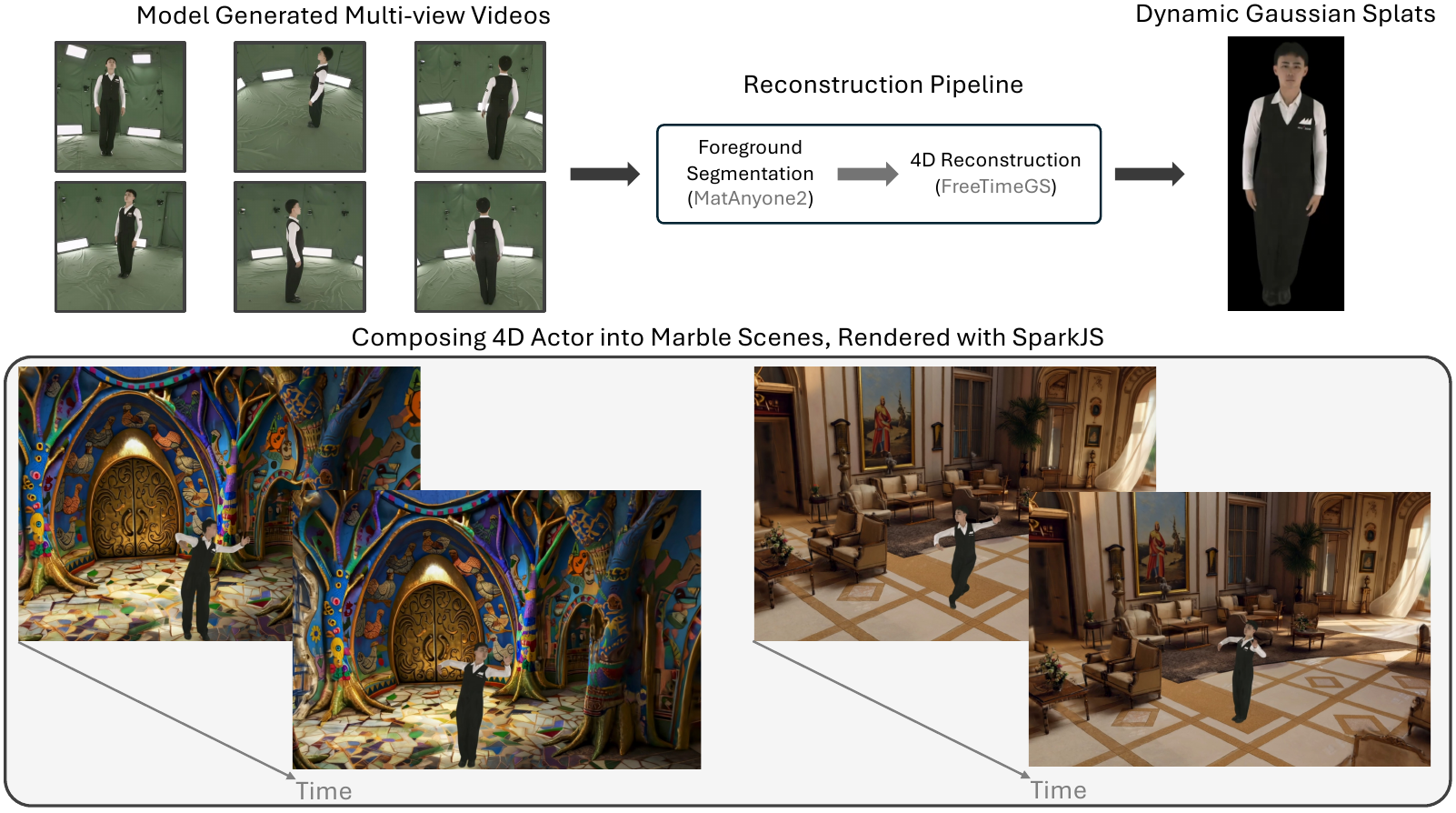}
    \caption{
    \textbf{Composing 4D actors into generated scenes.}
    Model-generated synchronized multi-view videos are segmented and reconstructed into dynamic Gaussian splats. The reconstructed
    actor is then composed into scenes and rendered in a browser environment, demonstrating a monocular-video-to-4D workflow
    for dynamic actors in generated worlds.
    }
  \label{fig:applications_scene_composition}
\end{figure*}

\modelname{} enables a practical
monocular-video-to-4D asset creation workflow. Given a monocular or sparse-view
actor video, our model first generates synchronized dense multi-view videos.
We then segment the foreground actor with MatAnyone2~\cite{matanyone2} and fit the generated views with
FreeTimeGS~\cite{ftgs} to reconstruct dynamic Gaussian splats. The resulting
4D actor can be composed into generated 3D scenes, such as Marble worlds~\cite{marble}, and
rendered interactively with SparkJS~\cite{sparkjs}. This demonstrates how
our generated multi-view videos can serve as an intermediate representation for
downstream 4D content creation, enabling applications such as AR/VR, gaming,
simulation, and video re-shooting.
\section{Limitations and Future Work}
\label{sec:limitations}

Our current training data is still dominated by static studio-capture rigs, limiting generalization to dynamic camera motion, in-the-wild environments, and highly out-of-distribution viewpoints such as extreme tilts or top-down views. While our PRoPE camera conditioning supports arbitrary per-frame camera transforms, robustness to these settings could be improved by incorporating dynamic multi-camera captures. 
Finally, although teacher-forced history conditioning enables temporal rollout, generating longer horizons requires additional drift-mitigation strategies, such as self-forcing~\cite{self_forcing}, diffusion forcing~\cite{diffusion_forcing}, or other long-horizon consistency objectives.
\section{Conclusion}
\label{sec:conclusion}

We presented \modelname{}, a multi-view video diffusion model that turns a
pre-trained text-to-video DiT into a generator of synchronized
multi-view videos from sparse reference views. Our method conditions
generation through a projective positional encoding that combines relative
camera geometry and permutation-invariant view encoding. Together with staged curriculum training and
multi-view appearance captions, this enables a single checkpoint to
generalize across varying view counts, camera layouts, and temporal
lengths without relying on explicit geometry priors.
Experiments on DNA-Rendering and ActorsHQ demonstrate state-of-the-art
performance in both in-distribution and zero-shot settings, outperforming
methods that depend on skeletons, depth, normals, or rendered target-view
geometry. 
We further show that the same formulation extends beyond humans to
multi-species animal generation and integrates directly with existing
4D Gaussian Splatting methods, enabling monocular-video-to-4D
applications such as AR/VR, gaming, simulation, and video re-shooting.~

\newpage
\begin{ack}
We thank Justin Johnson, Keunhong Park, Zixuan Huang, Justin Cui, Bardienus Pieter Duisterhof, Yi Hua, Karan Desai, Mohamed El Banani, Minhao Chen, and Raghav Garg for their valuable discussions. We also thank Christoph Lassner, Ben Mildenhall, and Fei-Fei Li for their support throughout this project.
We are grateful to Andreas Sundquist for his guidance on SparkJS, and to Brittani Poeppel and Ian Curtis for their assistance and guidance in the use of the Marble worlds.
\end{ack}

\clearpage
\bibliographystyle{plainnat}
\bibliography{references}

\clearpage

\end{document}